\documentclass[letterpaper]{article} 
\usepackage{aaai24}  
\usepackage{times}  
\usepackage{helvet}  
\usepackage{courier}  
\usepackage{algorithm}
\usepackage{algorithmic}
\usepackage{appendix}
\usepackage{amsmath}
\usepackage{threeparttable}
\usepackage{amsfonts}
\usepackage{mathrsfs}
\usepackage{booktabs}
\usepackage{graphicx} 
\usepackage{epstopdf}
\usepackage{subfigure}
\usepackage{multirow}
\usepackage{makecell}
\usepackage{caption}
\usepackage[hyphens]{url}  
\usepackage{graphicx} 
\usepackage{natbib}  
\usepackage{caption} 
\usepackage{algorithm}
\usepackage{algorithmic}
\usepackage{newfloat}
\usepackage{listings}
\DeclareCaptionStyle{ruled}{labelfont=normalfont,labelsep=colon,strut=off} 
\floatstyle{ruled}
\newfloat{listing}{tb}{lst}{}
\floatname{listing}{Listing}
\title{Object Attribute Matters in Visual Question Answering}
\author{
    Peize Li\textsuperscript{\rm 1}\equalcontrib,
    Qingyi Si\textsuperscript{\rm 2,3}\equalcontrib,
    Peng Fu\textsuperscript{\rm 2,3}\thanks{\rm Corresponding Authors. 
    },
    Zheng Lin\textsuperscript{\rm 2,3},
    Yan Wang\textsuperscript{\rm 1,4}\textsuperscript{\rm \dag}\\
    }
\affiliations{
    \textsuperscript{\rm 1}School of Artificial Intelligence, Jilin University, Changchun, China\\ 
    \textsuperscript{\rm 2}Institute of Information Engineering, Chinese Academy of Sciences, Beijing, China\\
    \textsuperscript{\rm 3}School of Cyber Security, University of Chinese Academy of Sciences, Beijing, China\\ 
    \textsuperscript{\rm 4}Key Laboratory of Symbol Computation and Knowledge Engineering of Ministry of Education, College of Computer Science and Technology, Jilin University, Changchun, China\\     

}

\usepackage{bibentry}

\begin{document}

\maketitle

\begin{abstract}
 Visual question answering is a multimodal task that requires the joint comprehension of visual and textual information. 
 However, integrating visual and textual semantics solely through attention layers is insufficient to comprehensively understand and align information from both modalities.  
 Intuitively, object attributes can naturally serve as a bridge to unify them,  which has been overlooked in previous research.
 In this paper, we propose a novel VQA approach from the perspective of utilizing  object attribute, 
 aiming to achieve better object-level visual-language alignment and multimodal scene understanding.
 Specifically, we design an attribute fusion module and a contrastive knowledge distillation module. 
 The attribute fusion module constructs a multimodal graph neural network 
 to fuse attributes and visual features through message passing. 
 The enhanced object-level visual features contribute to solving fine-grained problem like counting-question. 
 The better object-level visual-language alignment aids in understanding multimodal scenes, thereby improving the model's robustness. 
 Furthermore, to augment scene understanding and the out-of-distribution performance, 
 the  contrastive knowledge distillation module introduces a series of implicit  knowledge.
 We distill knowledge into attributes through  contrastive loss,
 which further strengthens the representation learning of attribute features and facilitates visual-linguistic alignment. 
 Intensive experiments on six datasets, COCO-QA, VQAv2, VQA-CPv2, VQA-CPv1, VQAvs and TDIUC, show the  superiority of the proposed method.
\end{abstract}

\section{Introduction}
\begin{figure}[t]
  \centering
  \includegraphics[width=3.3in]{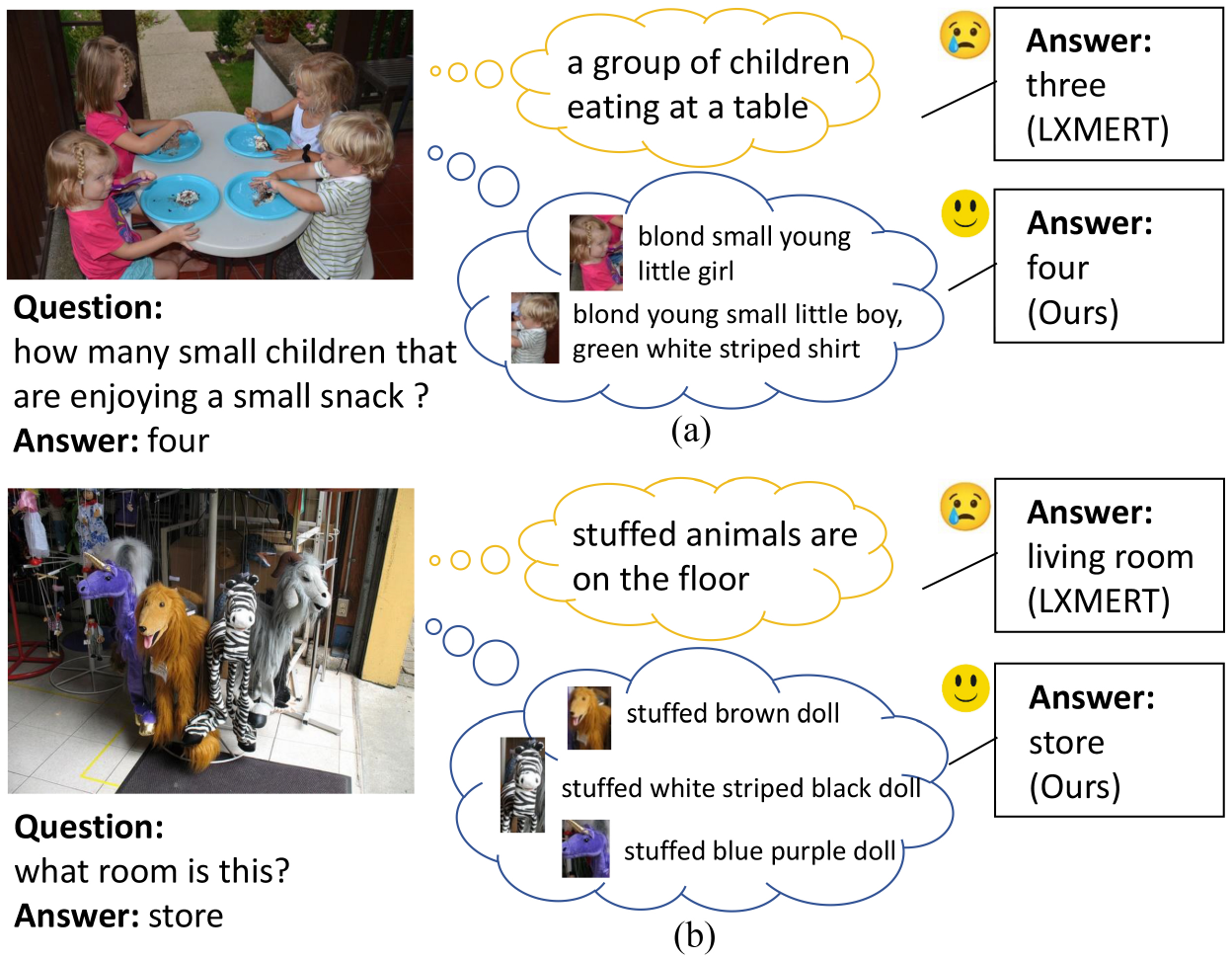}
  \caption{An illustration of our motivation. Compared with previous multimodal content, 
  object-level attributes are indispensable in both object counting (a) and scene understanding (b).
  }
\label{fig:motivation}
\end{figure}


Visual Question Answering is a multimodal task involving the interaction of vision and language, which aims to answer the question based on  visual image content. 
Most of the existing solutions \cite{kim2018bilinear, anderson2018bottom, li2019relation, yu2019deep, peng2020mra, si2023combo} depend on visual relations,  attention mechanisms and external knowledge to connect  question information and associated visual clues. 
Visual relations \cite{li2019relation, peng2020mra} provide semantic connections and relative positions between the objects, 
aiding in enhancing the spatial understanding of image content.
Attention mechanisms  \cite{kim2018bilinear, anderson2018bottom, yu2019deep} give co-occurrence information in multimodal scenes, 
which enables the model to concentrate on the important words and visual elements. 
External knowledge \cite{gao2022transform, gui2021kat} offers relevant background and topological relationships among the entities,
which contribute to understanding the contextual information  of multimodal scenes.
However, both of these lack attributes of visual objects, 
which can directly offer fine-grained semantic information about visual objects.
The object attributes cover a wide range of advanced concepts, including objects, scenes, actions and modifiers, 
which are indispensable for 
enhancing the understanding of object-level visual content 
and achieving object-level visual-language alignment.
Next, we can better  explain this idea through the following two examples.


An example from COCO-QA \cite{ren2015exploring} dataset is shown in Figure 1(a).
Answering this question first requires understanding the various types of children in the image and then calculating the number of children.
Object attributes provide different descriptive information for each child, 
which improves the model's ability to solve object-level fine-grained problem,
such as enhanced counting ability. 
Another example from VQA-CPv2 \cite{agrawal2018don} dataset is shown in Figure 1(b).
Answering this question requires combining the semantic information of multiple objects in the scene, 
and then the model makes a comprehensive judgment. 
Object attributes 
improve the model's ability to solve complicated 
scene-understanding problem, 
which boosts the out-of-domain (OOD) performance.
In summary, 
visual object attributes achieve object-level visual-language alignment, 
especially beneficial for the above two problems.
Therefore, \textbf{O}bject \textbf{A}ttribute \textbf{M}atters in \textbf{V}isual \textbf{Q}uestion \textbf{A}nswering (OAM-VQA).

Recently,  several methods have been well-developed to enhance VQA models using object attributes. 
Some prompt-based learning methods \cite{gui2021kat, gao2022transform, si2023combo} utilize attributes to design prompts, 
other methods \cite{agrawal2018don,anderson2018bottom,nguyen2022coarse} fuse attributes based on attention mechanisms.
However, none of them achieve strong object-level visual-language alignment.
As a result, they perform poorly  on  
object-level fine-grained problem
as well as complicated scene-understanding problem.


To address the aforementioned problem, 
we utilize object attributes to explicitly align visual and linguistic semantics. 
Specifically, our approach primarily consists of the Attribute Fusion Module (AFM) and the  Contrastive Knowledge Distillation  Module (CKDM).
Attribute Fusion Module establishs a novel multimodal graph neural network to fuse the visual features and object attributes.
Through updating nodes, 
the multimodal graph neural network iteratively aggregates information from neighboring nodes 
to capture detailed global information encompassing all objects.
This allows the Attribute Fusion Module to learn both the shared characteristics among all objects 
and their individual attributes.
In this way, the advanced object-level visual features 
contribute to addressing object-level fine-grained problem.

Contrastive Knowledge Distillation Module further enriches the representation of attribute features. 
Following TwO \cite{si2023combo}, 
this module firstly uses prompt to introduce a series of  implicit knowledge stored in the visual-language pre-trained
(VLP) models OFA \cite{wang2022ofa}, BLIP \cite{li2022blip} and BLIP2 \cite{li2023blip}.  
Then, it employs an enhanced transformer to encode knowledge.
Through contrastive loss,
we distill knowledge into attributes,
which enhances the understanding of scenes and  the model's robustness. 
Therefore, this module contributes to addressing complicated scene-understanding problem. 

In summary, this paper explores the role of object attributes in visual question answering, 
and finds that object attributes are beneficial for 
enhancing the understanding of object-level visual content 
and facilitating the alignment between object-level visual and linguistic elements. 
The main contributions of this work contain:
\begin{itemize}
\item We propose a novel and effective method OAM-VQA that leverages object attributes to explicitly unify the visual and linguistic semantics.   
\item We design an Attribute Fusion Module and a Contrastive Knowledge Distillation Module, which respectively contribute to addressing object-level fine-grained problem and complicated scene-understanding problem.
\item Extensive experimental results on six datasets, including COCO-QA, VQAv2, VQA-CPv2, VQA-CPv1, VQAvs and TDIUC,  validate the effectiveness and generality of our approach.
\end{itemize}

\section{Related Work} \label{Related Work}

\subsubsection{Incorporating Object Attribute in VQA.}
Recently, some inspiring works \cite{si2023combo, gui2021kat,gao2022transform,anderson2018bottom, nguyen2022coarse} attempt to incorporate object attributes to address the VQA task and achieve remarkable progress.  
UpDn \cite{anderson2018bottom} and VinVL \cite{zhang2021vinvl} 
directly leverage object attributes as input 
to learn effective visual representations.
Different from focusing on enhancing the object detector,
CFR-VQA \cite{nguyen2022coarse}  designs an elaborate BAN \cite{kim2018bilinear} 
to fuse attribute features. 
However, it also unavoidably introduces some noise or ambiguous attribute information. 
The prompt-based learning methods \cite{si2023combo, gui2021kat,gao2022transform} utilize attributes to obtain external knowledge from the knowledge base or VLP model. 
These methods excel in leveraging broader cross-domain knowledge to solve the VQA task. 
However, they fail to achieve object-level visual-language alignment, 
which could lack the capability to address 
object-level fine-grained problem and 
scene-level  understanding problem.
Our method goes further in both directions:
On the one hand, 
we establish the multimodal graph neural network to fuse object attributes.
On the other hand, we do not merely introduce  a series of knowledge.
Furthermore, 
we effectively utilize knowledge to enrich attribute feature representation and promote object-level visual-linguistic alignment.

In recent years, 
numerous studies \cite{si2022language, goyal2017making, ren2015exploring, si2023compressing} propose diverse VQA tasks
to evaluate different types of core skills
for addressing the visual question answering. 
One type of dataset focuses on image content understanding, 
such as COCO-QA \cite{ren2015exploring} and TDIUC \cite{kafle2017analysis}.
COCO-QA \cite{ren2015exploring} is automatically generated based on image captions 
and can be classified into four main types: object, color, number and location.
TDIUC \cite{kafle2017analysis} is a task-driven image understanding dataset,
where the questions can be categorized into 12 classes,
such as 
counting 
and sentiment understanding.
The OOD  datasets have a notable difference in answer distribution between the training and testing sets,  
and the models that only learn biases from the training set struggle to perform well on OOD datasets.
Common OOD datasets consist of VQA-CPv1/2  \cite{agrawal2018don}, VQAv2 \cite{goyal2017making} and VQAvs \cite{si2022language}.
They are proposed for studying language bias problem.
Furthermore, VQAvs \cite{si2022language} is a comprehensive dataset containing visual bias, language bias and multimodal bias. 
We validate our approach on multiple datasets, 
which cover two important settings: 
image content understanding and out-of-distribution robustness. 
In this way, 
the visual question answering ability of the model is comprehensively assessed.

\subsubsection{Graph Neural Network.}
Graph neural network (GNN) \cite{li2022dynamic, scarselli2008graph,li2019relation,gao2020multi, zhu2021mucko} is a highly effective framework for representing  graph-structured data.
GNNs follow the message passing scheme that updates each node’s feature using its neighborhoods of nodes to capture specific patterns of a graph. 
Some encouraging works \cite{li2022dynamic, li2019relation, gao2020multi, zhu2021mucko} study graph neural networks to solve the VQA task.
For example, ReGAT \cite{li2019relation} represents the image as a graph and captures interactions between objects through the graph attention mechanism. 
Moreover, Mucko \cite{zhu2021mucko} constructs a multimodal heterogeneous graph  consisting of visual features, image captions and factual knowledge.
It utilizes graph convolutional networks to capture multi-layer graph representations to predict the answers.
Unlike the aforementioned approaches that update nodes based on modality-specific information,
we establish a multimodal graph consisting of a visual sub-graph and an attribute sub-graph. 
Our approach updates node representation from interactions across different modalities to learn comprehensive attribute feature representations and better achieve object-level visual-linguistic alignment.



\section{Methodology}
\begin{figure*}[t]
  \begin{center}
  \includegraphics[width=5.0in]{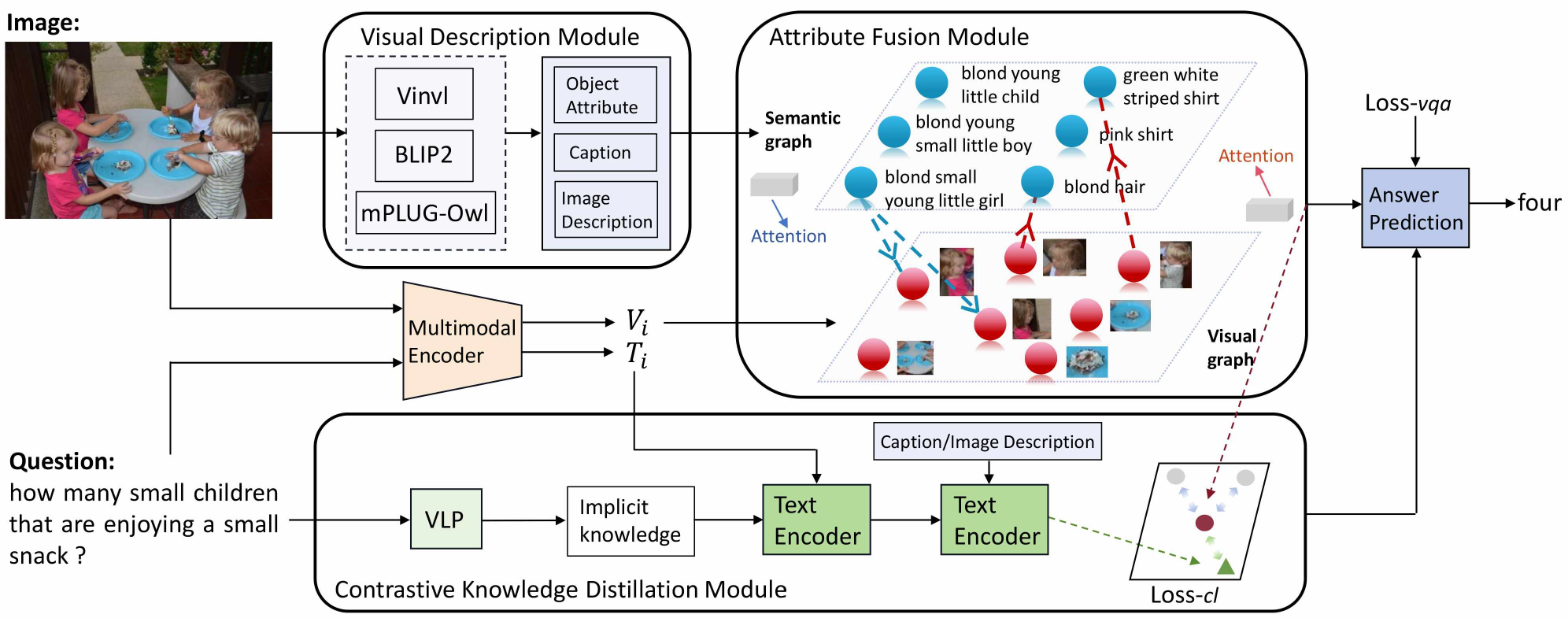}
  \end{center}
  \caption{
  The overview of our attribute-centric approach. Visual description module generates descriptive text for object attributes.
  Attribute fusion module establishes a multimodal graph 
  and fuses attribute features with visual features by passing messages between two subgraphs.
  Contrastive knowledge distillation module introduces a series of implicit knowledge to supplement information that cannot be covered in the attributes. 
  On this basis, the contrastive loss is adopted to further strengthen and enrich the representation of attribute features. 
  The blue or red arrows between nodes in the two graphs represent the direction of information flow. 
  }
\label{model}
\end{figure*}

In this section, we elaborate on the proposed OAM-VQA approach for visual question answering.
Figure \ref{model} shows OAM-VQA’s overview, which contains: 
multimodal encoding, 
visual description module, 
attribute fusion module,  
contrastive knowledge distillation module 
and answer prediction module.


\subsection{Multimodal Encoding}
A multimodal encoder is used to encode question and image features.
Most existing VQA models consider VQA as a multi-class classification task. 
Among them, LXMERT \cite{tan2019lxmert} is a transformer-based approach like BERT \cite{kenton2019bert},
and can encode visual objects and question text into visual features and textual features.
Besides, 
LXMERT is the most popular pre-trained visual-language model, 
thus we adopt it as the multimodal encoder in our proposed approach.
Concretely, given a VQA dataset 
$\mathcal{D}=\{(\boldsymbol{I}_i,\boldsymbol{Q}_i, \boldsymbol{A}_i)\}_{i=1}^{N}$ 
with $N$ samples, where $\boldsymbol{I}_i$,  $\boldsymbol{Q}_i$ and $\boldsymbol{A}_i$ are the image, question and ground-truth answer of $i$-th sample respectively. 
LXMERT  encodes image $\boldsymbol{I}_i$ and question $\boldsymbol{Q}_i$ separately in two streams, and extracts
visual features $\boldsymbol{V}_i =\{\boldsymbol{o}_{i1},\boldsymbol{o}_{i2},...,\boldsymbol{o}_{ij}\}_{j=1}^M$ 
and  question features $\boldsymbol{T}_i$. 
$M$  is the number of visual objects detected by Faster RCNN \cite{ren2015faster}. 
The visual features $\boldsymbol{V}_i$ will serve as the initial representation in the attribute fusion module, 
and question features $\boldsymbol{T}_i$ will be further utilized in the  contrastive knowledge distillation module.

\subsection{Visual Description Module}
Visual descriptions provide more descriptive semantic information about visual images,
which effectively reduces the semantic gap between the two modalities.
Given an image $\boldsymbol{I}_i$, following TwO \cite{si2023combo}, 
visual description module  generates descriptive text at different levels, 
consisting of object-level attributes, image-level  global captions and image-level detailed descriptions.
First, we use the VinVL detector \cite{zhang2021vinvl} to obtain object-level attributes and utilize 300-dimensional Glove \cite{pennington2014glove}  to acquire their word embeddings as the initial attribute features
$\boldsymbol{E}_{i} = \{\boldsymbol{e}_{i1},\boldsymbol{e}_{i2},...,\boldsymbol{e}_{ik}\}_{k=1}^L$. 
$L$ is the number of attribute words.
Then, we adopt the SOTA visual-language pretrained model BLIP2 \cite{li2023blip} to generate image-level  global  captions 
and obtain the corresponding encoded features $\boldsymbol{C}_{i}$ in the same way.
Besides, we apply a multimodal large language model mPLUG-Owl \cite{ye2023mplug} to generate image-level detailed descriptions. 
mPLUG-Owl \cite{ye2023mplug} is a multimodal model based on large language model. 
It has stronger language generation capabilities 
and is capable of generating descriptions more detailed than traditional image captions.
And in the ablation experiments, we compare it with object-level attributes and image-level captions to explore their effects in VQA.
The aforementioned  descriptions of visual content  will serve as the initial features for the subsequent attribute fusion module. 

\subsection{Attribute Fusion Module}
Attribute fusion module guides information passing between the visual graph and the semantic graph.
The goal of this module is to fuse object-level attributes and visual features to achieve better object-level visual-linguistic alignment.
\subsubsection{Multimodal graph construction.} Given an image $\boldsymbol{I}_i$, we first construct a  multimodal graph composed of two fully connected sub-graphs, i.e.,
visual graph $\boldsymbol{G}_{iv}$ and semantic graph $\boldsymbol{G}_{it}$ for representing two modalities of information.
In the visual graph $\boldsymbol{G}_{iv}$, 
each node $\boldsymbol{v}_{ij}$  
represents each visual object.
The initial representation $\boldsymbol{v}_{ij}^{(0)}$ is obtained through multimodal encoding.
We set $\boldsymbol{v}_{ij}^{(0)} = \boldsymbol{o}_{ij}$.
In the semantic graph $\boldsymbol{G}_{it}$, 
each node $\boldsymbol{s}_{ik}$ represents an object attribute. 
The initial node representation $\boldsymbol{s}_{ik}^{(0)}$
is the feature $\boldsymbol{e}_{ik} $ 
from visual description module.

\subsubsection{Aggregation scheme.}
After constructing multimodel graph and initializing the representation of each node, we propose two aggregators which guide the information flow between the visual graph and the semantic graph. 
This aggregation scheme leverages diverse types of contexts from different modalities to refine the node representations, as shown in Figure \ref{model}. 
The first aggregator utilizes attribute features to update the visual nodes. 
For each node $\boldsymbol{v}_{ij}$ in visual graph $\boldsymbol{G}_{iv}$,  the aggregator updates the representation of $\boldsymbol{v}_{ij}$ by attending on neighbour nodes in semantic graph  $\boldsymbol{G}_{it}$.
Concretely, we first calculate the relevance score between the node $\boldsymbol{v}_{ij}$ and its neighboring node $\boldsymbol{s}_{ik}$ as below:
\begin{equation} \label{eq1}
 \boldsymbol{r}_{s_{ik}, v_{ij}}^{'} = f_v(\boldsymbol{v}_{ij}^{(0)})^{T}(f_s(\boldsymbol{s}_{ik}^{(0)}))
\end{equation}
\begin{equation} \label{eq2}
 \boldsymbol{r}_{s_{ik}, v_{ij}}^{v} = \frac{\exp{({\boldsymbol{r}}_{s_{ik}, v_{ij}}^{'}})}{{\sum\limits_{s_{ik} \in \mathcal{N}_{v_{ij} ^ {s}}} \exp{(\boldsymbol{r}_{s_{ik}, v_{ij}}^{'}})}}
\end{equation}
where $f_s$ and $f_v$ are the multi-layer perceptron (MLPs) that used to encode node features. 
$\mathcal{N}_{v_{ij} ^ {s}}$ indicates the neighboring nodes of
$v_{ij}$ in  the semantic graph. 
After that, we aggregate the information of attended nodes from the semantic graph to the visual graph.
Each visual node is updated: 
\begin{equation} \label{eq3}
 \boldsymbol{v}_{ij} ^{(1)} = [\boldsymbol{v}_{ij} ^{(0)}; \sum\limits_{s_{ik} \in \mathcal{N}_{v_{ij} ^ {s}}}\boldsymbol{r}_{s_{ik},v_{ij}}^{v} f_{s^{'}}(\boldsymbol{s}_{ik} ^{(0)})]
 \small
\end{equation}
where 
$f_{s^{'}}$ is an MLP to encode the features of neighboring nodes from the semantic graph.
$[;]$ denotes the concatenation of two vectors.
Similar to the update mechanism for visual nodes, 
we further obtain the updated attribute representations as follows:

\begin{equation} \label{eq4}
 \boldsymbol{r}_{s_{ik}, v_{ij}}^{s} = \frac{\exp{({\boldsymbol{r}}_{s_{ik}, v_{ij}}^{'}})}{{\sum\limits_{v_{ij} \in \mathcal{N}_{s_{ik} ^ {v}}} \exp{(\boldsymbol{r}_{s_{ik}, v_{ij}}^{'}})}}
\end{equation}

\begin{equation} \label{eq5}
 \boldsymbol{s}_{ik} ^{(1)} = [\boldsymbol{s}_{ik} ^{(0)}; \sum\limits_{v_{ij} \in \mathcal{N}_{s_{ik} ^ {v}}}\boldsymbol{r}_{s_{ik},v_{ij}}^{s} f_{v^{'}}(\boldsymbol{v}_{ij} ^{(1)})]
\end{equation}
where $f_{v^{'}}$ is an MLP used to encode the features of neighboring nodes. 
$\mathcal{N}_{s_{ik} ^ {v}}$ represents the neighboring nodes of
$s_{ik}$ in  the visual graph.

\subsection{Contrastive Knowledge Distillation Module}
Contrastive knowledge distillation module aims to further consolidate the representation learning of attribute features.
Firstly, 
we introduce a series of implicit knowledge, 
and then distill knowledge into attributes through contrastive loss.
This further enhances the  understanding of scenes, 
greatly boosting OOD robustness.
\subsubsection{Text encoding.}
Specifically, inspired by prompting GPT-3, we first utilize prompts to acquire implicit 
knowledge stored in  VLP models OFA, BLIP and BLIP2. 
We adopt the question $\boldsymbol{Q}_i$ and the image $\boldsymbol{I}_i$ as prompt to generate exploratory answers  and obtain its word embeddings
$\boldsymbol{P}_i$.

To better encourage the alignment between tokens, 
following compound token attention \cite{aladago2022compound}, 
we adopt an enhanced transformer method based on channel fusion to encode features.
It maps the question features $\boldsymbol{T}_i$ and implicit
knowledge features $\boldsymbol{P}_i$ separately into half of the embedding space:
\begin{equation} \label{eq6}
\boldsymbol{P}_i ^{'}=f_1(\boldsymbol{P}_i)
 \small
\end{equation}
\begin{equation} \label{eq7}
\boldsymbol{T}_i ^{'}=f_2(\boldsymbol{T}_i)
 \small
\end{equation}
where $f_1$ and $f_2$ are MLPs. 
Subsequently, we employ two cross-attention layers to independently encode the features 
and then merge the original features:
\begin{equation} \label{eq8}
\boldsymbol{\hat T}_i=[\boldsymbol{T}_i ^{'};H_1(\boldsymbol{T}_i ^{'}, \boldsymbol{P}_i ^{'}, \boldsymbol{P}_i ^{'})]
 \small
\end{equation}
\begin{equation} \label{eq9}
\boldsymbol{\hat P}_i=[\boldsymbol{P}_i ^{'};H_2(\boldsymbol{P}_i ^{'}, \boldsymbol{T}_i ^{'}, \boldsymbol{T}_i ^{'})]
 \small
\end{equation}
where $H_1(q,k,v), H_2(q,k,v)$ are two cross-attention function with $q$, $k$, and $v$ as query, key and value respectively. 
Then, it combines the features of the two input streams to create compound tokens and learns the final representation through a self-attention function $G_{att}(x)$:

\begin{equation} \label{eq10}
\boldsymbol{Z}_i=G_{att}([\boldsymbol{\hat T}_i;\boldsymbol{\hat P}_i])
 \small
\end{equation}
As a result, we obtain the fused features $\boldsymbol{Z}_i$ of the question feature $\boldsymbol{T}_i$ and  implicit 
knowledge 
$\boldsymbol{P}_i$. 
This encoding process can adequately focus on question-related knowledge from implicit knowledge.
Next, we further fuse the obtained $\boldsymbol{Z}_i$ with image caption $\boldsymbol{C}_i$
to acquire the representation of image-related parts. 
Consequently, we acquire the encoded knowledge feature $\boldsymbol{F}_i$. 

Finally,  we utilize two top-down attention networks \cite{anderson2018bottom} to obtain question-oriented attribute features and  knowledge features, 
formulated as,
\begin{equation} \label{eq11}
\boldsymbol{\overline S}_i=f_{att} ^{s}(\boldsymbol{S}_i, \boldsymbol{Z}_i)^{T} \boldsymbol{S}_i
 \small
\end{equation}

\begin{equation} \label{eq12}
\boldsymbol{\overline F}_i=f_{att} ^{t}(\boldsymbol{F}_i, \boldsymbol{Z}_i)^{T} \boldsymbol{F}_i 
 \small
\end{equation}
where $f_{att} ^{s}$ and $f_{att} ^{t}$ are top-down attention networks, $\boldsymbol{Z}_i$ is the question feature after transformer encoding.

\subsubsection{Contrastive loss.} 
Inspired by the LRC-BERT method \cite{fu2021lrc},
which employs contrastive learning for latent semantic distillation in the intermediate layers, 
we use contrastive loss to distill knowledge into attributes.
%
Given question-related attribute features $\boldsymbol{\overline S}_i$ and  knowledge features $\boldsymbol{\overline F}_i$,  
we construct positive sample pairs $(\boldsymbol{\overline S}_i, \boldsymbol{\overline F}_i^{+})$  and negative sample pairs  $(\boldsymbol{\overline S}_b, \boldsymbol{\overline F}_b)_{b=1}^{B}$ in the same batch. ($b \neq i$). 
$B$ is the number of negative samples in a batch. 
Following MMBS \cite{si2022towards}, we adopt  the cosine similarity as the scoring function. 
The contrastive loss is formulated as:
\begin{equation} \label{eq15}
L_{cl}= -\log\frac{e^{cos(\boldsymbol{\overline S}_i, \boldsymbol{\overline F}_i^{+})}}{e^{cos(\boldsymbol{\overline S}_i, \boldsymbol{\overline F}_i^{+})}+\sum\limits_{b=1}^B {e^{cos(\boldsymbol{\overline S}_b, \boldsymbol{\overline F}_b)}}}
 \small
\end{equation}

\subsection{Answer Prediction Module}
The answer prediction module takes the question-oriented attribute features $\boldsymbol{\overline S}_i$ and  knowledge features $\boldsymbol{\overline F}_i$ as inputs, and outputs the answer, 
as follows: 
\begin{equation} \label{eq13}
{\boldsymbol Y}_{i} ^{pre} =f_{ans} ([\boldsymbol{\overline S}_i;\boldsymbol{\overline F}_i])
 \small
\end{equation}
where $f_{ans}$ represents an MLP used to calculate the scores for all candidate answers.
The overall training objective comprises two components: the VQA multi-label classification loss $L_{vqa}$ and the contrastive loss $L_{cl}$. 

\section{Experiments}
\begin{table}[t]
  \centering
  \resizebox{1.0\columnwidth}{!}{
    \begin{tabular}{lcccc}
    \toprule
    Dataset & \#QA pairs  & \#Images & Image Source\\
    \midrule
    COCO-QA & 118K  & 123K  & COCO  \\
    TDIUC & 1.6M  & 167K  & COCO + VG \\
    VQA-CPv1 & 370K  & 205K  & COCO \\
    VQA-CPv2 & 603K  & 219K  & COCO \\
    VQAv2 & 1.1M  & 204K  & COCO \\
    VQAvs & 658K  & 877K  & COCO \\    
    \bottomrule
    \end{tabular}%
    }
    \caption{Comparison of datasets used in this paper. VG represents Visual Genome dataset.}
  \label{data}%
  \vspace{-1em}
\end{table}%

\subsection{Dataset and Experimental Settings}
\subsubsection{Dataset.}
We assess the performance of our approach on image understanding datasets (COCO-QA, TDIUC) and OOD datasets (VQA-CPv1, VQA-CPv2, VQAv2 and VQAvs),
which validates its capability in addressing image-understanding problem and OOD problem respectively.
The dataset statistics can be found in Table \ref{data}.
For the detailed introduction to the datasets, 
please refer to Related Work.
\subsubsection{Experimental settings.}
Our model is trained by AdamW optimizer with 100 epochs. 
The self-attention function $G_{att}(x)$ in the   module consists of 5 layers of self-attention.
In both the cross-attention and self-attention layers, the hidden layer dimension is 512,  and the number of heads is  8.

\subsection{Comparisons with State-of-the-Arts}
\subsubsection{Comparison on image understanding datasets.} 
\begin{table}[t]
  \centering
  \resizebox{1.0\columnwidth}{!}{
    \begin{tabular}{l|c|cccc}
    \hline
     Methods  & All   & Objects & Number & Color & Location \\
    \hline
    SAN (\citeyear{yang2016stacked})   & 61.60  & 65.40  & 48.60  & 57.90  & 54.00  \\
    QRU (\citeyear{li2016visual})  & 62.50  & 65.06  & 46.90  & 60.50  & 56.99  \\
    HieCoAtt (\citeyear{lu2016hierarchical}) & 65.40  & 68.00  & 51.00  & 62.90  & 58.80  \\
    Dual-MFA (\citeyear{lu2018co}) & 66.49  & 68.86  & 51.32  & 65.89  & 58.92  \\
    CVA (\citeyear{song2018pixels}) & 67.51  & 69.55  & 50.76  & 68.96  & 59.93  \\
    MCAN (\citeyear{yu2019deep})  & 68.08  & 69.39  & 54.19  & 71.52  & 60.17  \\
    ODA  (\citeyear{wu2018object})  & 69.33  & 70.48  & 54.70  & 74.17  & 60.90  \\
    CoR (\citeyear{wu2018chain})  & 69.38  & 70.42  & 55.83  & 74.13  & 60.57  \\
    CAM (\citeyear{peng2022answer})  & 69.68  & 70.32  & 55.26  & 77.10  & 59.28  \\
    ALSA (\citeyear{liu2022alsa}) & 69.97  & 71.59  & 54.83  & 72.74  & 61.78  \\
    MCAN+PA (\citeyear{mao2022positional}) & 70.10  & 71.13  & 55.97  & 74.85  & 62.07  \\
    MRA-Net (\citeyear{peng2020mra}) & 70.27  & 71.40  & 56.42  & 74.69  & 60.62  \\
    \hline
    \textbf{OAM-VQA} & \textbf{75.22}  & \textbf{75.67}  & \textbf{68.20}  & \textbf{80.66}  & \textbf{63.80}  \\
    \hline
    \end{tabular}%
    }
    \caption{Comparison with the state-of-the-art approaches on the COCO-QA dataset.}
  \label{COCO}%
\end{table}%

We compare our method with the state-of-the-art methods on \textbf{COCO-QA} dataset in Table \ref{COCO}. 
Our proposed method consistently outperforms the state-of-the-art MRA-Net with comfortable margin (70.27$\%$ $\sim$ 75.22$\%$ absolute accuracy
improvement).
In particular, OAM-VQA improves performance (from 56.42$\%$ $\sim$  to 68.20$\%$) on number-questions.
MRA-Net \cite{peng2020mra} explores various visual  relationships to improve model performance, 
while we bring in object attributes that provide more and deeper visual semantics. 
This directs the model's focus more towards the objects themselves, 
\textit{thereby enhancing its ability to handle counting-type questions.}
In Table \ref{VQA-CPv1}, we evaluate our model on the \textbf{TDIUC} dataset. 
The results show that our method achieves the highest performance, specifically surpassing MuRel 2.42$\%$.
These findings indicate that \textit{our approach  utilizes object attributes to enhance the understanding of visual content, 
thus excelling in solving object-level fine-grained questions.}
\begin{table}[t]
  \centering
  \resizebox{1.0\columnwidth}{!}{
    \begin{tabular}{lc|lc}
    \hline
    Methods & TDIUC & Methods & VQA-CP v1 \\
    \hline
    MCB (\citeyear{fukui2016multimodal}) &81.86  & SAN (\citeyear{yang2016stacked})  & 26.88  \\
    SAN (\citeyear{yang2016stacked})  & 82.00  & NMN (\citeyear{andreas2016neural}) & 29.64  \\
    RAU (\citeyear{kafle2017analysis})   & 84.26 & MCB (\citeyear{fukui2016multimodal})   & 34.39  \\
    QTA (\citeyear{shi2018question})   & 85.03  & Counter (\citeyear{zhang2018learning}) &  37.67 \\
    BAN (\citeyear{kim2018bilinear})  & 85.50  & GVQA (\citeyear{agrawal2018don})  & 39.23  \\
    DFAF (\citeyear{gao2019dynamic}) & 85.55  & UpDn (\citeyear{anderson2018bottom}) & 39.74  \\
    BLOCK (\citeyear{ben2019block}) & 85.96  & LXMERT$\dagger$ (\citeyear{tan2019lxmert}) & 52.21  \\
    CoR (\citeyear{wu2018chain})   & 86.91  & AdvReg (\citeyear{ramakrishnan2018overcoming}) & 43.43  \\
    MIRTT (\citeyear{wang2021mirtt}) & 87.50  & RUBi (\citeyear{cadene2019rubi}) & 50.90  \\
    MLI (\citeyear{gao2019multi})  & 87.60  & LMH (\citeyear{clark2019don})   & 55.73  \\
    MRA-Net (\citeyear{peng2020mra}) & 87.73  & CCS+UpDn (\citeyear{chen2020counterfactual})   & 60.95  \\
    DCAF (\citeyear{liu2019densely}) & 88.0   & AdaVQA+UpDn (\citeyear{adaVQA}) & 61.20  \\
    MuRel (\citeyear{cadene2019murel}) & 88.20  & CL (\citeyear{liang2020learning})   & 61.27  \\
    \hline
    \textbf{OAM-VQA} & \textbf{90.62}  & \textbf{OAM-VQA}  & \textbf{65.43}  \\
    \hline
    \end{tabular}%
    }
    \caption{Comparison with the state-of-the-art approaches on the TDIUC and VQA-CPv1 datasets.}
  \label{VQA-CPv1}%
\end{table}%


\begin{table}[t]
  \centering
  \resizebox{1.0\columnwidth}{!}{
    \begin{tabular}{lcccc|cccc}
    \hline
    \multicolumn{1}{c}{\multirow{2}{*}{Methods}} & \multicolumn{4}{c|}{VQA-CP v2 test} & \multicolumn{4}{c}{VQAv2 val} \\
\cline{2-9}          & \multicolumn{1}{l}{All} & \multicolumn{1}{l}{Y/N} & \multicolumn{1}{l}{Num} & \multicolumn{1}{l|}{Other} & \multicolumn{1}{l}{All} & \multicolumn{1}{l}{Y/N} & \multicolumn{1}{l}{Num} & \multicolumn{1}{l}{Other} \\
    \hline
    \textbf{Base models} &  &  &  &   &   &  &  &  \\
    SAN (\citeyear{yang2016stacked})  & 24.96  &38.35  &11.14  & 21.74  & 52.41  &70.06  &39.28  &47.84  \\
    BAN   (\citeyear{kim2018bilinear}) & 37.03 & 41.55 & 12.43 & 41.4  & 63.9  & 81.42 & 45.18 & 55.54 \\
    UpDn (\citeyear{anderson2018bottom})  & 39.74 & 42.27 & 11.93 & 46.05 & 63.48 & 81.18 & 42.14 & 55.66 \\
    LXMERT$\dagger$ & 51.85 & 54.38  & 26.92  & 58.01  & 70.94 & 87.92  & 57.57  & 61.33  \\
    \hline
    \textbf{Debiasing methods} &  &  &  &   &   &  &  &  \\ 
    AdvReg (\citeyear{ramakrishnan2018overcoming})  & 41.17  &65.49  & 15.48  &35.48  & 62.75  & 79.84  &42.35 &55.16  \\
    HINT (\citeyear{selvaraju2019taking}) & 46.73  &70.04  & 10.68  & 46.31  & 63.38  &81.18  & 42.99  &55.56  \\
    RUBi (\citeyear{cadene2019rubi}) &47.11  &68.65  & 20.28  & 43.18  &61.16  &-  &-  &-  \\
    SCR (\citeyear{wu2019self})& 48.47 & 70.41 & 10.42 & 47.29 & 62.30  & 77.40  & 40.90  & 56.50  \\
    LMH (\citeyear{clark2019don}) & 52.45 & 69.81 & 44.46 & 45.54 & 61.64 & 77.85 & 40.03 & 55.04 \\
    CF-VQA (\citeyear{niu2021counterfactual}) & 53.55 & 91.15 & 13.03 & 44.97 & 63.54 & 82.51 & 43.96 & 54.3 \\
    MMBS$\ast$ (\citeyear{si2022towards}) & 56.51 & 79.83 & 28.70  & 51.92 & 70.85 & 88.25 & 55.67 & 61.63 \\
    CSS  (\citeyear{chen2020counterfactual}) & 58.95 & 84.37 & 49.42 & 48.21 & 59.91 & 73.25 & 39.77 & 55.11 \\
    Re-scaling$\ast$ (\citeyear{guo2021loss}) & 66.40  &79.77  &59.06  &61.41  &69.76   &85.32  &52.07  &62.60  \\    
    SAR$\ast$ (\citeyear{si2021check}) & 66.73 & 86.00  & 62.34 & 57.84 & 69.22 & 87.46 & 51.20  & 60.12 \\
    MUTANT$\ast$ (\citeyear{gokhale2020mutant}) & 69.52 & 93.15 & 67.17 & 57.78 & 70.24 & 89.01 & 54.21 & 59.96 \\
    MDDC$\ast$ (\citeyear{li2023multi}) & 69.77 & 87.88 & 52.8  & 64.93 & 74.51 & 90.14 & 58.81 & 66.76 \\
    \hline
    \textbf{OAM-VQA} & 60.97 & 69.98  & 53.09  & 58.41  & 71.99 & 88.63  & 57.94  & 63.25  \\
    \hline
    \end{tabular}%
    }
    \caption{Comparison with the state-of-the-art approaches on the VQA-CPv2 test and VQAv2 val datasets. $\dagger$ denotes our implementation. $\ast$ indicates that the models adopt LXMERT as the baseline.}
  \label{VQA-CPv2}%
\end{table}%

\begin{table}[t]
  \centering
  \resizebox{1.0\columnwidth}{!}{
    \begin{tabular}{lccc|c}
    \hline
    Methods & \makecell[c]{Language-\\bias} & \makecell[c]{Visual- \\ bias} & \makecell[c]{Multimodal-\\ bias} & Average \\
    \hline
    S-MRL (\citeyear{cadene2019rubi})& 43.03  & 31.65  & 49.48  & 42.65  \\
    UpDn (\citeyear{anderson2018bottom}) & 47.22  & 37.35  & 52.55  & 46.80  \\
     + LMH (\citeyear{clark2019don}) & 46.33  & 37.56  & 50.75  & 45.85  \\
     + LMH-L (\citeyear{clark2019don})& 47.33  & 36.08  & 52.38  & 46.51  \\
     + LMH-V (\citeyear{clark2019don})& 46.68  & 36.93  & 52.28  & 46.38  \\
     + SSL (\citeyear{zhu2021overcoming})& 45.98  & 36.43  & 51.28  & 45.62  \\
    BAN  (\citeyear{kim2018bilinear}) & 48.97  & 38.51  & 54.65  & 48.53  \\
    \hline
    LXMERT$\dagger$ & 53.13  & 41.17  & 61.05  & 53.16  \\
    \textbf{OAM-VQA} & \textbf{53.87}  & \textbf{42.10}  & \textbf{61.23}  & \textbf{53.71}  \\
    \hline
    \end{tabular}%
    }
    \caption{Comparison with the state-of-the-art approaches on the VQAvs dataset. For example, language bias contains keyword bias, visual bias consists of key object bias, and multimodal bias involves combinations of the two.}
  \label{VQAvs}%
\end{table}%


\begin{figure}[t]
    \centering
    \begin{center}
    \includegraphics[width=3.2in]{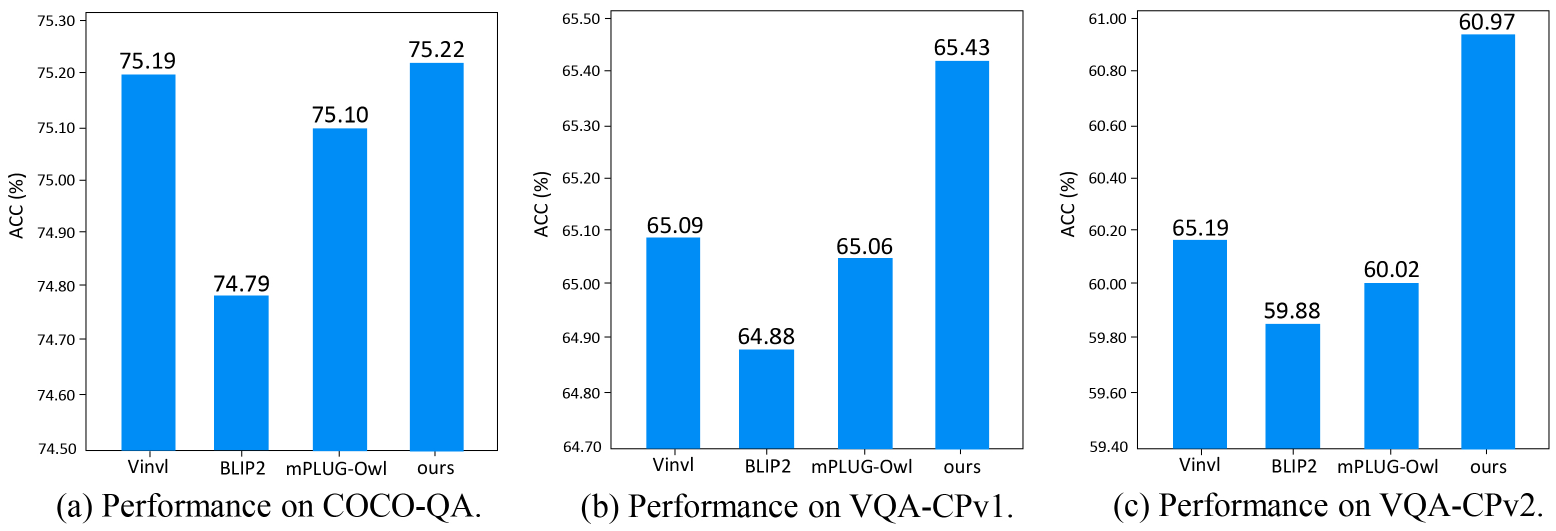}
    \end{center}
    \vspace{-1em}
    \caption{Performance with different types of visual descriptions. Vinvl generates object-level attributes, BLIP2 generates image-level global captions and mPLUG-Owl generates image-level detailed descriptions.}
    \label{fig:h_vis}
\end{figure}

\begin{table}[t]
  \centering
  \resizebox{1.0\columnwidth}{!}{
    \begin{tabular}{l|ccc}
    \hline
    \multicolumn{1}{c}{\multirow{2}[4]{*}{Models}} & \multicolumn{3}{c}{Datasets} \\
\cmidrule{2-4}    \multicolumn{1}{c}{} & \multicolumn{1}{c}{COCO-QA} & \multicolumn{1}{c}{VQA-CPv1} & \multicolumn{1}{c}{VQA-CPv2} \\
    \hline
    LXMERT & 72.61 & 52.21 & 51.85 \\
    LXMERT+CKDM &74.94  &63.97  &58.33  \\
    LXMERT+AFM & 75.19 & 65.09 & 60.19 \\
    LXMERT+AFM+CKDM (ours) & \textbf{75.22} & \textbf{65.43} & \textbf{60.97} \\
    \hline
    \end{tabular}%
    }
    \caption{Ablation of key components in OAM-VQA on COCO-QA, VQA-CPv1 and VQA-CPv2. ``AFM" represents Attribute Fusion Module, and ``CKDM"  stands for Contrastive Knowledge Distillation Module.}
  \label{tab:addlabe4}%
\end{table}%

\subsubsection{Comparison on OOD datasets.}

Table \ref{VQA-CPv2} shows the comparison results on the \textbf{VQA-CPv2} test.
Unlike the datasets mentioned above, 
the plain VQA models without debiasing methods perform poorly on these biased datasets. 
Therefore, we compare our model with plain VQA models and debiasing methods.
Brief descriptions of baseline models are in Appendix A. 
For the VQA-CPv2 test, our approach improves the backbone LXMERT with a large performance gain (+9.12$\%$).
Specifically, on the number-questions, 
our model achieves a 26.17$\%$ boost.
In Table \ref{VQA-CPv1}, 
our approach outperforms the CL method by 4.16$\%$ on the \textbf{VQA-CPv1} dataset.
Existing debiasing methods for VQA-CP often rely on  its construction characteristic that ``the answer distribution under the same question type in the training set and test set are almost reverse'' \cite{si2022language,teney2020value}. 
Therefore, 
the latest SOTA methods like MDDC, SAR and MUTANT can always perform best.  
In contrast, our method does not use such dataset-specific characteristic and also achieves competitive performance. 
Besides, most debiasing methods tend to enhance the performance of VQA-CP at the expense of sacrificing the performance of VQAv2 (e.g., CSS, LMH, SAR),
while our approach achieves improvements on both datasets, showing genuine out-of-distribution robustness. 
We also achieve favorable performance on the VQAv2 dataset presented in  Table \ref{VQA-CPv2}, surpassing LXMERT by 1.05$\%$.
Table \ref{VQAvs} displays the performance for alleviating the language biases, visual biases and multimodal biases on \textbf{VQAvs}.
In terms of language bias and visual bias, 
our model outperforms LXMERT by 0.74$\%$ and 0.93$\%$ respectively. 
These results demonstrate that \textit{our approach leverages object attributes to enhance the understanding of scenes, 
thereby boosting the OOD performance.}

\subsection{Ablation Study}
We conduct ablation studies on the COCO-QA, VQA-CPv1 and VQA-CPv2 datasets to
examine the effectiveness of our approach.  
COCO-QA serves as a representative dataset for image understanding, 
VQA-CPv1/v2 represent out-of-distribution (OOD) datasets. 
From Figure \ref{fig:h_vis} and Table \ref{tab:addlabe4}, several observations can be derived:
(1) In the Figure \ref{fig:h_vis}, we assess the effectiveness of different levels of descriptive text about visual content.
We find that the model with object attributes  performs the best. 
This is because object-level visual-linguistic alignment is more effective than global alignment. 
In addition, the performance gains brought by image captions are slightly higher than those of image descriptions.
(2) In Table \ref{tab:addlabe4}, we study the ablation of key components of our method. 
We observe that the attribute fusion module achieves comparative improvements 
(+2.58$\%$ on COCO-QA, +12.88$\%$ on VQA-CPv1 and +8.34$\%$ on VQA-CPv2) 
compared to LXMERT. 
This is because the attribute fusion module effectively fuses object attribute with visual features
through a multimodal graph neural network.
Besides, 
we notice that the contrastive knowledge distillation module further enhances the performance. 
This is because this module introduces a series of textual knowledge to further enrich the representation of attribute features
and promotes visual-linguistic alignment through contrastive loss.
Furthermore, we investigate the impact of different types of knowledge on datasets in Appendix B. 
We find that implicit knowledge from OFA contributes the most to OOD datasets. 
The knowledge from BLIP2 has a greater impact on the image understanding datasets. 
Although both BLIP2 and OFA are visual language pre-training models with encoder-decoder structure, 
the decoder in BLIP2 is a large language model.
\textit{Containing more visual information in the question helps to stimulate more knowledge from the large language model.}
Therefore, for the image understanding dataset COCO-QA, 
BLIP2 offers more efficient knowledge.
More detailed examples are shown in Appendix B.


\subsection{Analysis}
\subsubsection{Performance on different question types.}

\begin{figure}[t]
    \centering
    \begin{center}
    \includegraphics[width=3.2in]{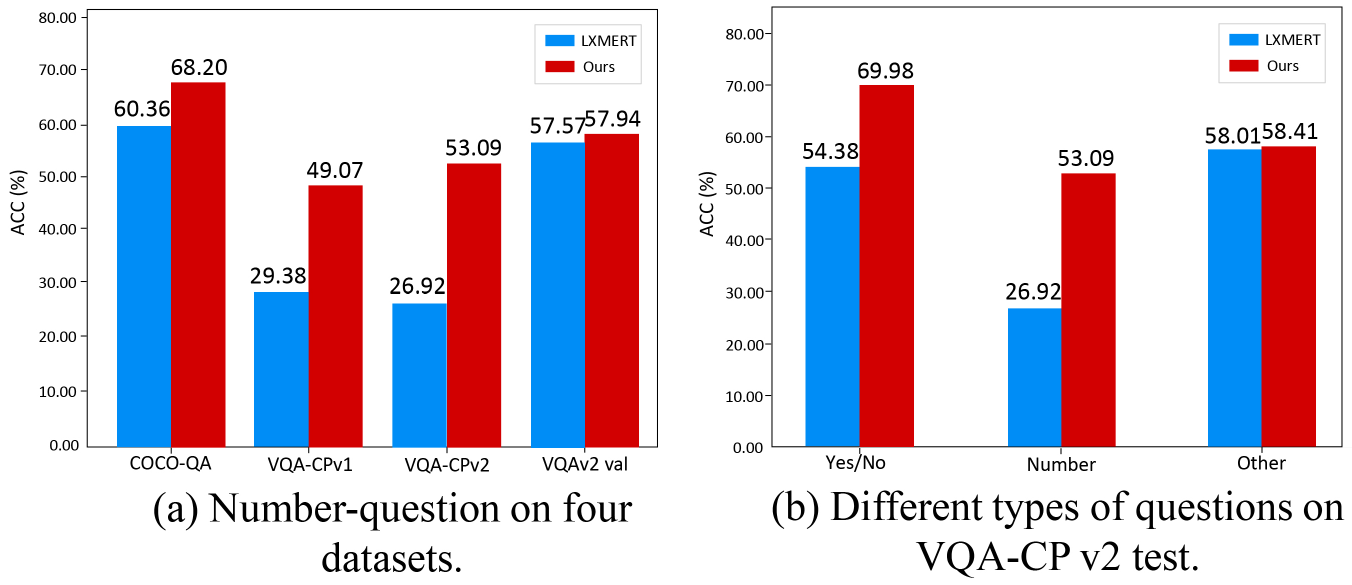}
    \end{center}
    \vspace{-1em}
    \caption{Performance with different question types. The red bar represents our approach, and the blue bar represents LXMERT.}
    \label{fig:3}
\end{figure}

From Table \ref{COCO} and Table \ref{VQA-CPv2}, 
we investigate the comparison between our approach and LXMERT across different question types, 
including object, number, color, location, yes/no and other questions. 
For number questions, our method achieves remarkable improvements of 7.84$\%$, 26.17$\%$ and 0.37$\%$ on the COCO-QA, VQA-CPv2 and VQAv2 datasets respectively compared to LXMERT. 
Regarding yes/no questions, our method   outperforms LXMERT by  +15.60$\%$ on VQA-CPv2 and +0.71$\%$ on VQAv2 dataset.
This also supports our conclusion: 
\textit{object attributes enhance object-level visual understanding, 
aiding in addressing object-level fine-grained  problem}. 

In Figure \ref{fig:3}, 
we further visualize the performance comparison of our approach and LXMERT across different question categories.
From Figure 4(a), 
it is evident that our approach significantly outperforms LXMERT on number-question across all four datasets. 
In Figure 4(b), for the VQA-CPv2 dataset, 
our approach outperforms LXMERT by 15.60$\%$, 26.17$\%$ and 0.40$\%$ on Yes/No, number and other questions respectively. 
This  result demonstrates that \textit{our method excels not only in number-question but also remains highly effective  across a broader range of question types and datasets}.
Therefore, we conclude that object attribute matters in visual  question answering.



\subsubsection{Qualitative analysis.}

\begin{figure}[t]
    \centering
    \begin{center}
    \includegraphics[width=3.3in]{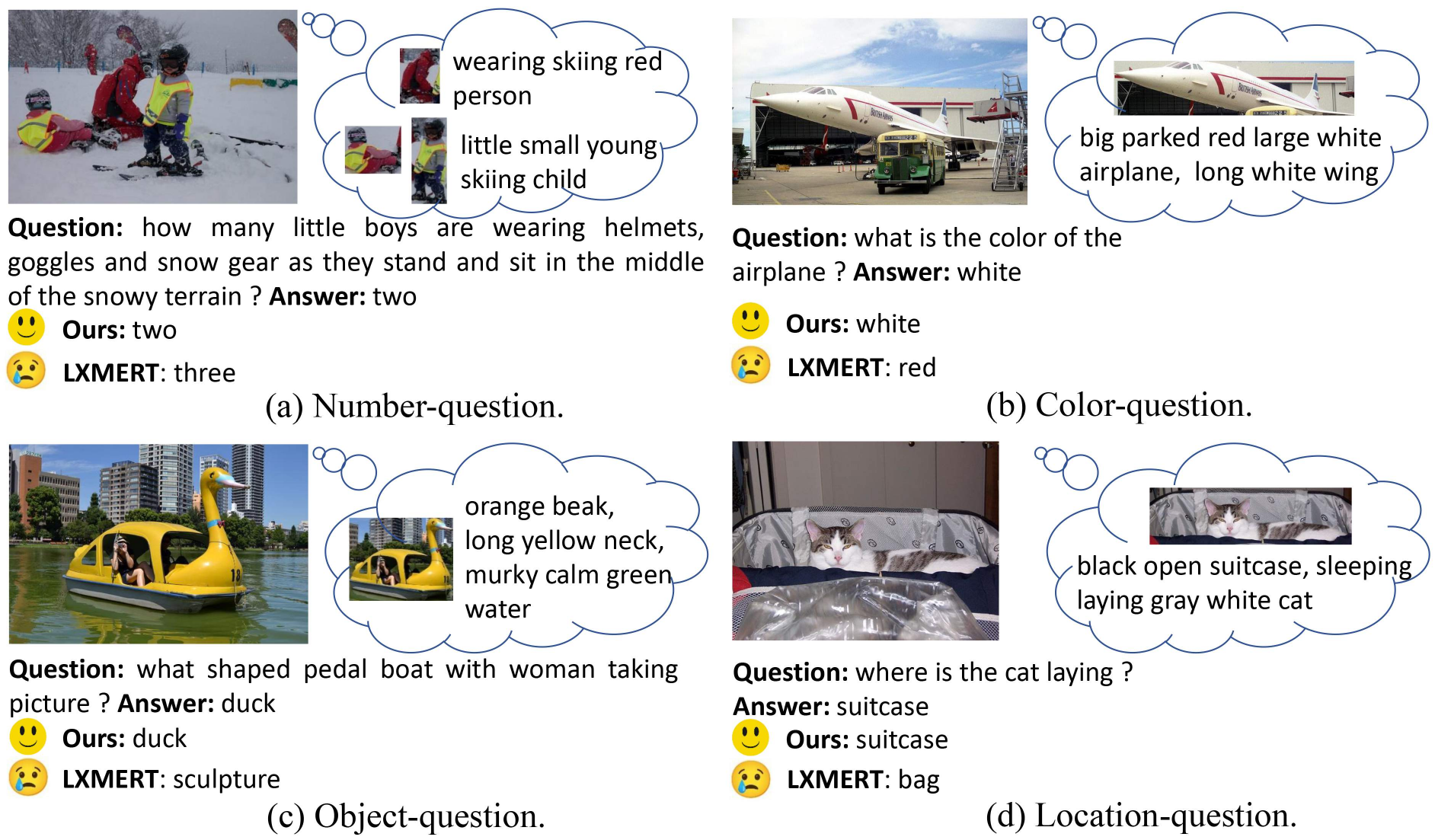}    
    \end{center}
    \vspace{-1em}
    \caption{Examples of four different question types on the COCO-QA dataset.}
    \label{fig:5}
\end{figure}

In Figure \ref{fig:5}, we analyze examples from four question types on the COCO-QA dataset: number, color, object and location. 
We conclude the following two insights:
(1) In multimodal scenarios with noise interference, 
object attributes enable the model to pay greater attention to
question-oriented visual objects. 
In Figure 5(a), there are some children and an adult. 
When calculating the number of children, 
the adult adds complexity to the model. 
However, 
our approach uses the attribute fusion module to fuse object attributes and visual features, 
thereby enhancing the understanding of visual content. 
Our approach overcomes those interferences and answers the question correctly.
In Figure 5(b), 
the noise is the red rectangular box on the white airplane. 
The object attributes provide more descriptive information about the airplane,
and help our method overcome the noise and understand the overall color of the airplane.
However, LXMERT lacks these object-level fine-grained attributes and answers the question incorrectly.
(2) For complex scene-understanding questions, 
object attributes offer valuable answer-related clues.
In Figure 5(c) and Figure 5(d), 
we see that the object attributes provide relevant information about the correct answer, 
such as orange beak, long yellow neck, and black open suitcase.
Our attribute-centric approach 
effectively fuses these attributes
and 
answers these questions correctly.

\section{Conclusion}
In this paper, we propose an effective method to achieve object-level visual-linguistic alignment.
Our method designs an attribute fusion module to fuse object attributes with visual features, 
thus enhancing the understanding of object-level visual content. 
Subsequently, 
through the contrastive knowledge distillation  module,
we introduce a series of implicit knowledge from visual-language pre-trained model, 
further reinforcing the representation learning of attribute features.
Through contrastive loss, 
we distill knowledge into attributes. This further  enhances the understanding of scenes and greatly improves the OOD performance.
Extensive experiments conducted on image understanding datasets (COCO-QA and TDIUC) and OOD datasets (VQA-CPv1/v2, VQAv2 val and VQAvs) demonstrate the advantages of our approach.
We explore the role of describing visual content text from different levels. 
We hope that our work will encourage more attention to the understanding of object attribute, promoting the advancement of VQA.

\section{Acknowledgments}
This work was supported by the National Natural Science Foundation of China (Nos.  62072212, 61976207), the Development
Project of Jilin Province of China (Nos. 20220508125RC, 20230201065GX), and the Jilin Provincial Key Laboratory of
Big Data Intelligent Cognition (No. 20210504003GH). 


\bibliography{aaai24}

\end{document}